# Adversarial examples attack based on random warm restart mechanism and improved Nesterov momentum

Tiangang Li
Wuhan University
Wuhan, Hubei, China
zaiwuhan2014@gmail.com

## ABSTRACT

The deep learning algorithm has achieved great success in the field of computer vision, but some studies have pointed out that the deep learning model is vulnerable to attacks adversarial examples and makes false decisions. This challenges the further development of deep learning, and urges researchers to pay more attention to the relationship between adversarial examples attacks and deep learning security. This work focuses on adversarial examples, optimizes the generation of adversarial examples from the view of adversarial robustness, takes the perturbations added in adversarial examples as the optimization parameter. We propose RWR-NM-PGD attack algorithm based on random warm restart mechanism and improved Nesterov momentum from the view of gradient optimization. The algorithm introduces improved Nesterov momentum, using its characteristics of accelerating convergence and improving gradient update direction in optimization algorithm to accelerate the generation of adversarial examples. In addition, the random warm restart mechanism is used for optimization, and the projected gradient descent algorithm is used to limit the range of the generated perturbations in each warm restart, which can obtain better attack effect. Experiments on two public datasets show that the algorithm proposed in this work can improve the success rate of attacking deep learning models without extra time cost. Compared with the benchmark attack method, the algorithm proposed in this work can achieve better attack success rate for both normal training model and defense model. Our method has average attack success rate of 46.3077%, which is 27.19% higher than I-FGSM and 9.27% higher than PGD. The attack results in 13 defense models show that the attack algorithm proposed in this work is superior to the benchmark algorithm in attack universality and transferability.

## CCS CONCEPTS

• Insert CCS text here • Insert CCS text here • Insert CCS text here

## 1 INTRODUCTION

The concept of adversarial examples was first proposed by the research team represented by Szegey[1]. From the view of optimization, they found that when there is a slight perturbation in the image, it will affect the classification results. Subsequently, Ian Goodfellow[2] and others improved this optimization method, further perfected the concept of adversarial examples, and proposed regularization constraints based on optimal maximum norm. The attack algorithm using gradient information propagated by neural network to disturb original examples is named fast gradient sign method (FGSM), which can directly and efficiently generate adversarial examples. Kurakin[3, 4] et al. improved FGSM's one-step direct method in fixed direction, and used BIM, a basic iterative method that iterated multi-step and accumulated small-amplitude perturbations to generate adversarial examples. Moosavi[5] and others proposed an optimization-based DeepFool algorithm to attack classification decision plane, which greatly improves the success rate of adversarial attacks compared with the earliest optimization algorithm. Moosavi[6] et al. proposed general adversarial examples through experiments, which is generated by generating training set with the same distribution of sampling subsets. The general perturbations do not pay attention to original examples, but only focuses on the attacked models. The perturbations can be uniformly added to datasets to attack the models in batches.

In this work, the adversarial robustness of neural network is studied from the view of model adversarial robustness optimization. The adversarial perturbations is regarded as a parameter that needs to be optimized, and the parameters of the attacked model are fixed. According to the optimization algorithm based on Nesterov momentum[7] used in the training process of deep learning model, the characteristics of fast convergence and ideal optimization results are positive. Combined with the idea of limiting the feasible region of gradient in Projected Gradient Descent, PGD[8] attack algorithm based on improved Nesterov momentum is proposed. Under the condition of ensuring the runtime, it can generate the adversarial examples with higher attack success rate. This work mainly has the following three contributions:

1. Original Nesterov momentum is improved, and the number of back propagation needed to calculate gradient is saved. Nesterov momentum is introduced into the gradient-based iterative optimization algorithm, which can accelerate the convergence speed compared with the standard optimization algorithm. Accumulating previous gradient to update parameters can better avoid falling into local optima; The approximate second-order property of Nesterov momentum can improve the update direction of gradient in time and obtain better optimization results;

2. After researching the previous work, it is found that dividing the adding process of perturbations into multiple iterative processes will increase the computation and time cost, but it increase the attack success rate of adversarial examples. Moreover, when the generated adversarial examples are initialized, adding randomly initialized adversarial perturbations to the normal examples also improve the attack success rate. Based on the above two findings, this work proposes random warm restart[9] mechanism to optimize the adversarial examples. In each warm restart, the update step size of adversarial perturbations is adjusted according to cosine annealing;

3. Improved Nesterov momentum and random warm restart mechanism proposed in this work can be combined with all adversarial examples generation algorithms, which are based on gradient optimization to improve attack success rate.

## 2 PRELIMINARIES

### 2.1 Robustness optimization of adversarial

In this work, with the view of neural network's robustness and concept of gradient optimization[10], we take image classification model based on neural network as the research object, a more effective adversarial examples generation algorithm is proposed. Firstly, the normal examples inputted into model are defined as $x$, the corresponding true label is $y$. The functional form of classification model is defined as $f_\theta(x)$, in which $\theta$ represents a set of model parameters. The loss function of model is defined as $L(f_\theta(x), y)$, this is the loss function of normal training, and the target is to optimize the parameters and continuously reduce the loss function $L(f_\theta(x), y)$. Next, define adversarial examples of normal examples as $x^{adv}$, the adversarial perturbations added to normal examples is $\delta$, the perturbations is added as follow:

$$x^{adv} = x + \delta \quad (1)$$

Therefore, the target of generating adversarial examples is to search for suitable $\delta$ in the restricted space. The perturbations are added to normal examples, constantly increasing adversarial loss function $L(f_\theta(x^{adv}), y)$ [11]. Until the classification model outputs false results. The evaluation metrics of adversarial perturbations mainly uses $L_1$, $L_2$ and $L_\infty$ norm of perturbations. We adopt $L_\infty$ norm in this work. Define an allowable set of perturbations as $\Delta$ and the magnitude of perturbation is $\epsilon$, which are specifically expressed as follow:

$$\Delta = \{\delta : \|\delta\|_\infty \leq \epsilon\} \quad (2)$$

Finally, we introduce the target of adversarial robustness optimization: min-max problem, which is expressed as below. $D$ Represent a set of finite approximate examples to approximate the true examples distribution:

$$minimize_\theta \frac{1}{|D|} \sum_{x,y \in D} max_{\|\delta\| \leq \epsilon} L(f_\theta(x+\delta), y) \quad (3)$$

The above formula represents the optimization process of the adversarial robustness, and the inner maximization represents continuous optimization of adversarial perturbations $\delta$. Increasing the loss function misleads the classification error of model within limited perturbations range. Outer minimization means continuous optimization of model parameters $\theta$. And the loss function can be reduced to ensure that model can also classify adversarial examples correctly. Because of the non-convex nature of neural network, the problems of inner maximization and outer minimization cannot be solved globally. This work focuses on how to generate effective adversarial examples to solve the problem of inner maximization.

### 2.2 Related work

**FGSM.** The calculation process of FGSM is as follows. $x$ indicates normal examples without adding perturbations, $y$ indicates the classification result of model, $\delta$ indicates the adversarial perturbations added to normal examples, and attackers need to control that the slight perturbations cannot be found after being added to original examples. The adding form of perturbations is shown in formula (1). Adversarial examples are inputted into classification model, $x$ multiplies parameter matrix $w^T$, which is shown as follow:

$$w^T x^{adv} + b = w^T x + w^T \delta + b \quad (4)$$

The generation formula of adversarial perturbations is shown in formula (5). The gradient direction of parameter update is the change direction of original examples, and appropriate perturbations is added in this direction.

$$\delta = \epsilon \cdot sign(w) \text{ s.t. } \|\delta\|_\infty < \epsilon \quad (5)$$

Sign function can ensure that the adding direction of perturbations in original examples is same as the gradient change direction propagated by loss function. Then, for a model with higher dimensional parameters, as long as a slight perturbation is added in the same dimensional direction as gradient change direction, a perturbation which can affect the output result of model can be generated as a whole. After the perturbations are processed by model, they will have a great influence on final classification result, which leads to the deviation of output classification result.

Above is the case that model is linear. For deep learning model with complex network structure and huge weight parameters, although it can be regarded as a highly nonlinear model, the learning ability of model will be limited because of nonlinear elements. Partial activation functions such as sigmoid, tanh and ReLU have linear characteristics, which will reduce the nonlinearity of model, as well as the linear full connection layer before model output. Deep learning model cannot defend against adversarial perturbations caused by adversarial examples. In deep learning model, you can regard $w$ as the derivative of loss function with respect to input $x$, which is as shown as follow:

$$w = \nabla_x L(f_\theta(x), y) \quad (6)$$

The generation of perturbations $\delta$ is shown in formula (7), and the size of perturbations is determined by the basic perturbations limit $\epsilon$, which means the infinite norm of perturbations is less than

$\varepsilon$. It ensures that perturbations added in adversarial examples are too small to be detected by human.

$$\delta = \varepsilon \cdot sign(\nabla_x L(f_\theta(x), y)) \text{ s.t. } \|\delta\|_\infty < \epsilon \quad (7)$$

**I-FGSM.** FGSM only carries out once single gradient update, once the attack direction is determined, it can't be changed. The success rate is low when the attack is adversarial examples, especially when there is a target-oriented attack. Therefore, an iterative FGSM attack algorithm is proposed based on single FGSM attack, which can add slight perturbations to the original image several times. After each iteration, the pixel value of overflow threshold is clipped to control perturbations in the reasonable range. $\alpha$ represents step size in each iteration. The formula is shown as follows:

$$x_0^{adv} = x \quad (8)$$
$$x_{N+1}^{adv} = Clip_{x,\varepsilon}\{x_N^{adv} + \alpha \cdot sign(\nabla_x L(x_N^{adv}, y))\} \qu(9)$$

**MI-FGSM.** MI-FGSM[12, 13] refers to momentum iterative FGSM. It is also an iterative attack algorithm based on gradient. The concept of momentum is introduced. In conventional optimization algorithm, momentum can accelerate the convergence of loss function and make update direction more stable. It can also avoid the oscillation of optimization process due to gradient update and loss function falling into saddle point or local optimal solution in the optimization process. In each iteration of MI-FGSM, the added perturbations are not only related to current gradient, but also affected by the accumulated gradient calculated previously. The expression of adversarial examples generation is shown as follows:

$$x_0^{adv} = x, g_0 = 0 \quad (10)$$
$$g_{t+1} = \mu \cdot g_t + \frac{\nabla_x L(x_t^{adv}, y)}{\|\nabla_x L(x_t^{adv}, y)\|_1} \quad (11)$$
$$x_{t+1}^{adv} = lip(x_t^{adv} + \alpha \cdot sign(g_{t+1})) \quad (12)$$

The above process repeats multi-iteration, while limiting $\|x^{adv} - x\|_\infty \leq \varepsilon$ to define step size $\alpha = \varepsilon/T$ of each iteration attack. And $\mu$ is the decay coefficient of accumulated momentum. The algorithm can not only carry out effective white box attack, but also complete black box attack well. And it can be extended to target attack and infinite norm constraint attack[14].

**PGD.** PGD is projected gradient descent, which belongs to the category of gradient based iterative attack and can be regarded as multi-step FGSM-based. FGSM only updates once along the specified direction, while PGD adopts training mode of multiple cycles. Each cycle only needs a small amount of update, and next cycle will change the direction of perturbations generation. And regularization method is used to limit the maximum perturbations by projected gradient method. Different from previous I-FGSM, PGD uses gradient projected to limit perturbations exceeding the threshold. If perturbations are beyond the high-dimensional space whose radius is $\epsilon$ in each iteration, perturbations are mapped back to the threshold hyperplane with radius $\epsilon$. PGD can ensure that perturbations are not too large. It is convenient to iterate for many times to find the optimal point. The formulas is shown as follows:

$$x_{t+1} = \prod_{x+\Delta}(x_t + \alpha g(x_t)/\|g(x_t)\|_2) \quad (13)$$
$$g(x_t) = \nabla_x L(f_\theta(x_t), y) \quad (14)$$

The constraint space of perturbations is defined in formula (2), $\alpha$ indicates step size of each iteration.

## 3 METHODOLOGY

### 3.1 Nesterov momentum optimization

In the training of deep learning model, optimization of model parameters is the key to determine the model performance[15]. Optimization problem refers to finding a set of parameters $\theta$ in neural network to reduce loss function $L(f_\theta(x), y)$. The loss function includes performance evaluation and regular terms of model in train datasets. The most basic optimization algorithm is gradient descent. In deep learning model, because of the existence of local optima and saddle points, the optimized objective function may have many local optima that are not globally optimal, and saddle points with gradient values close to 0. Especially when the dimension of search space rises, it becomes more difficult to optimize loss function to find the optimal value[16]. Therefore, stochastic gradient descent[17], which is widely used in deep learning model training, is proposed. Although this algorithm has some improvement: sampling several examples from datasets independently and identically to calculate mean gradient value as an unbiased estimation of global gradient, and learning rate is updated according to the change of loss function with time. However, it's still difficult to overcome the problems of falling into local optima and slow convergence[18].

In order to solve above problems, an optimization algorithm based on momentum is proposed, which accumulates the moving average of the previous gradient exponential decay and continues to move along the direction after the average. From the formula, the momentum-based optimization algorithm introduces variables $v$ as velocity. It represents the direction and velocity of parameter moving in parameter space, and the velocity is set as exponential decay average of back gradient. The velocity of gradient contribution decay is determined by $\mu$, which is usually between 0 and 1. Learning rate is $\epsilon$. The parameter updating rules of momentum-based optimization algorithm are shown as follows:

$$v_{t+1} = \mu v_t - \epsilon \nabla f(\theta_t) \quad (15)$$
$$\theta_{t+1} = \theta_t + v_{t+1} \quad (16)$$

Momentum-based algorithm updates a part of the update amount of the previous iteration every time the parameters are updated, so as to smooth the gradient of current iteration. Figure 1 shows the gradient calculation process based on momentum. It can be seen that the current gradient and the momentum accumulated before are calculated in the form of vector to obtain a new gradient, which participates in the current parameter update of model. The new gradient corrects the direction and amplitude of original gradient.

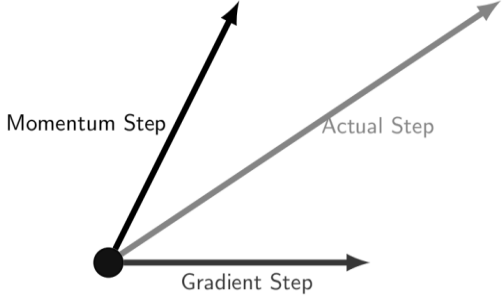

**Figure 1.** Momentum-based gradient calculation process

The optimization algorithm based on Nesterov momentum is a variant of the standard momentum optimization algorithm. Compared with the standard momentum method, Nesterov momentum adds a controllable parameter to introduce more momentum information. The parameter updating rules of the optimization algorithm based on Nesterov momentum are shown as follows:

$$v_{t+1} = \mu v_t - \epsilon \nabla f(\theta_t + \mu v_t) \quad (17)$$
$$\theta_{t+1} = \theta_t + v_{t+1} \quad (18)$$

Parameters $\mu$ and $\epsilon$ in the formula is similar to that of standard momentum method mentioned above. The difference between them lies in the calculation of gradient. When calculating Nesterov momentum, the part about gradient is calculated after increasing current velocity. Figure 2 shows the calculation process of Nesterov momentum. Combined with the standard momentum formula, it can be found that in Nesterov momentum calculation, Gradient is not calculated directly based on current parameters $\theta_t$, but according to the parameters obtained after taking the step originally planned. And the gradient is calculated by $\theta_t + \mu v_t$.

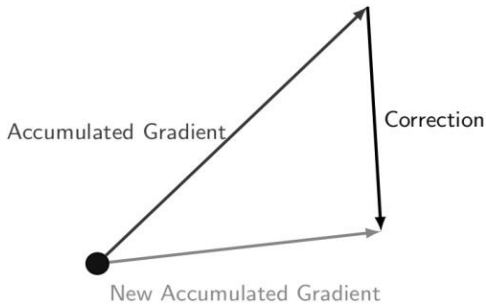

**Figure 2.** Nesterov momentum-based gradient calculation process

### 3.2 Improved Nesterov momentum optimization

By observing standard form of Nesterov momentum, it is found that every time the parameters are updated, the gradient must be recalculated and the gradient accumulated in previous iterations is not fully utilized. It will increase the calculation amount. Meanwhile, in order to explain that the Nesterov momentum-based optimization algorithm converges faster than the standard momentum-based optimization algorithm, the formula of Nesterov momentum algorithm is transformed The equivalent forms are shown as follows:

$$v_{t+1} = \mu v_t - \epsilon \nabla f(\theta_t) + \mu\epsilon[\nabla f(\theta_{t-1}) - \nabla f(\theta_t)] \quad (19)$$
$$\theta_{t+1} = \theta_t + v_{t+1} \quad (20)$$

By comparing difference between Nesterov equivalent formula and standard momentum formula, when Nesterov momentum is updated once, it will increase $\mu\varepsilon[\nabla f(\theta_{t-1}) - \nabla f(\theta_t)]$ in this update direction. It is equivalent to one more change of current gradient relative to last gradient. When current gradient is larger than last gradient, it is speculated that it will continue to increase and the part that is speculated to increase is calculated in advance. If the gradient is smaller than last time, the reduced part should be subtracted in advance. The added term approximates second derivative of the loss function. The mathematical essence of Nesterov momentum-based is to increase second derivative information of the loss function, which can converge faster.

Nesterov momentum, which is similar to the second order, contains more information of gradient change, but the Nesterov momentum-based optimization algorithm is not as strict on convergence conditions as Newton method and other second order methods, and its calculation is simpler. The Nesterov momentum-based optimization algorithm is essentially first-order algorithm with low computational complexity. According to original Nesterov momentum and the actual calculation requirements of attack algorithm proposed in this work, the original calculation formula is transformed equivalently, which mainly omits the inverse derivation after updating parameters in formula (17). In the formula (19) after equivalent transformation[19], only current gradient and previous gradient are needed for parameter update of next iteration. And there is no need to carry out back derivation after parameter update. This improvement reduces computational cost and is more convenient for implementation of the algorithm. The attack algorithm described below is implemented based on the equivalent form mentioned above.

### 3.3 Random warm restart mechanism

Based on stochastic gradient descent algorithm with warm restart in neural network training, this work proposes a random warm restart mechanism in adversarial examples generation. In the generation of adversarial examples, the hyperparameter step size $lr$ is very important in each iteration. Gradient can indicate the update direction of adversarial perturbations, and the value of each update is determined by step size. In this work, the attack algorithm is divided into multiple cycles of restart[20]. When each cycle is restarted, the input examples are adversarial examples that returned in previous cycle, and the step size is reinitialized. Then step size changes according to certain rules. In theory, the initial step size can be set to a larger value, and subsequent step size can be appropriately decayed. This method of periodically changing the step size can avoid falling into local optimal in the process of finding adversarial perturbations. In this work, random warm restart mechanism is combined with cosine annealing method, which is used as periodic function. The specific expression is shown as follow:

$$lr_t = lr_{min}^i + \frac{1}{2}(lr_{max}^i - lr_{min}^i)\left(1 + cos\left(\frac{T_{cur}}{T_i}\pi\right)\right) \quad (21)$$

Where $lr_{min}^i$ and $lr_{max}^i$ respectively represent minimum value and maximum value of step size; $T_{cur}$ indicates the iterations of attack algorithm from beginning to end in each restart, and step size changes with iterations in each restart; $T_i$ indicates the number of iterations for each restart.

### 3.4 Improved Nesterov momentum-based PGD

In this work, from the view of gradient optimization, the improved Nesterov momentum is introduced to generate perturbations and solve the inner maximization in robustness min-max problem. The adversarial perturbations are regarded as the only optimization parameter of Nesterov momentum optimization algorithm[21]. When initializing adversarial examples, random noise is added to the examples as initial adversarial examples perturbations. In the attack algorithm proposed, the projection operation is based on the projected gradient descent algorithm for each generated adversarial example and the projection operation is expressed as follows:

$$\delta := P(\delta + \alpha \nabla_\delta L(f_\theta(x+\delta), y)) \quad (22)$$

Where $P$ denotes the projection onto the ball of interest, in our work, we clip the perturbations in the case of $L_\infty$ norm.

The algorithm proposed introduces Nesterov momentum in the gradient update stage of standard PGD algorithm, and introduces random warm restart mechanism in generating perturbations. In each restart, we adjust step size of each iteration in NM-PGD based on cosine annealing method, and we run NM-PGD multiple times from different random locations within the $L_\infty$ ball. Because there are local optima that PGD started at the zero point will find, and which can be avoided to some extent just by randomization. The improved Nesterov momentum attack algorithm is shown in Algorithm 1, and the NM-PGD attack algorithm based on random warm restart mechanism is shown in Algorithm 2. The detailed algorithms for our attack methods are shown in Appendix A.

## 4 EVALUATION

In this work, two experiments are set up to evaluate the success rate of the algorithm proposed. For normally trained models, there normal convolutional network and ResNet-18. For defense model[22, 23, 24], there are adversarially trained models[25, 26, 27] and other advanced defense models[28, 29], which are totally 13 models. The attack methods are all white-box no target attacks, which can obtain the logits output of model and gradient. Attack results is $L_\infty$ norm of adversarial perturbations[30].

### 4.1 Experimental setup

**Datasets.** MNIST and CIFAR-10 are selected as the datasets used in our experiments. In experiment 1, all test datasets are used, which are 10,000 pictures. In experiment 2, the first 1000 pictures of CIFAR-10 test dataset are selected.

**Targeted models.** In experiment 1, a three-layer convolutional neural network model is trained on MNIST with classification accuracy of this model reaching 98.31%; the ResNet-18 model is trained on CIFAR-10 with classification accuracy reaching 93.58%. In experiment 2, 13 defense models trained on CIFAR-10 are selected. The attack tests of these models are completed based on ARES (Advanced Robustness Evaluation for Safety), ARES is a Python library for adversarial machine learning research focusing on benchmarking adversarial robustness on image classification correctly and comprehensively. These defense models not only include models based on adversarial training, such as cifar10-pgd_at based on PGD attack, cifar10-free_at based on recycling the gradient information, cifar10-fast_at combined with FGSM and randomly initialized, cifar10-at_he. Besides some other advanced defense models are included.

**Baselines.** We choose the benchmark algorithm FGSM, I-FGSM and PGD for experiments, showing the performance improvement of RWR-NM-PGD algorithm proposed in this work.

**Evaluation Metrics.** In essence, the attack process of adversarial examples is to generate examples, and then input the examples into targeted models for classification. The attack performance of the algorithm is evaluated by error rate of model classification of adversarial examples, which means the attack success rate of algorithm. The calculation formula is as follows:

$$Rate = \frac{1}{|M|}\sum_{M_i \in M} \frac{1}{|X|}\sum_{(x_j, y_j) \in X} 1\left(M_i\left(attack(x_j)\right) \neq y_j\right) \quad (23)$$

Where $M$ represent a set of all targeted models, $X$ is test dataset. The distance between all generated adversarial examples and original examples, which is the adversarial perturbations, is less than $\epsilon$.

**Hyperparameter setting.** The step size of FGSM in Experiment 1 is $\epsilon$. The iteration number of PGD is 5, and step size is $\epsilon/5$. The step size of RWR-NM-PGD is adjusted according to cosine function. For testing the attack results under different upper limits of perturbations, adversarial perturbations $\epsilon$ is 0, 0.05, 0.1, 0.15, 0.2, 0.25 and 0.3 respectively. The limitation of adversarial perturbations in experiment 2 is $8/255$. For each adversarial example, the average gradient calculation times are less than 100 times, and the average model prediction times are less than 200 times.

### 4.2 Attack normal models

Experiment 1 is the test of attacking normal model, which mainly evaluates the success rate and the generation time of adversarial examples.

*4.2.1 Attack success rate.* For MNIST and CIFAR10, the attack results of examples generation algorithm are as follows. The horizontal axis represents different perturbations restriction conditions, and the vertical axis represents attack success rate. Compared with FGSM and PGD on two datasets, RWR-NM-PGD proposed in this work has the best attack success rate on both target models. Experiments show that RWR-NM-PGD can achieve 100% attack success rate by properly adjusting the perturbations limit on both datasets, and it is superior to the other algorithms under any adversarial perturbations limit. As shown in Table 1 and Figure 3, when the perturbations limit is 0.3, RWR-NM-PGD has the success rate of 100%. As shown in Table 2 and

Figure 4, the perturbations limit is 0.15, and the attack success rate can reach 100%.

Further analysis results show that PGD and RWR-NM-PGD are much superior to FGSM under different perturbations restrictions, which also proves that the iterative training method will take more time to some extent, but the attack success rate is also higher. Meanwhile, the less the perturbations limit is, the higher the success rate of our method. This means our method performs better than baselines especially under strict attack conditions. Experimental results show that our method is very effective to attack standard convolutional neural network and classical CNN model ResNet-18 on public datasets, but there are still some problems.

**Table 1.** Attack success rate of adversarial attacks against CNN model on MNIST. Comparison of FGSM, PGD and RWR-NM-PGD

| MNIST $\epsilon$ | Success rate | | |
|---|---|---|---|
| | FGSM | PGD | RWR-NM-PGD(Ours) |
| 0.05 | 0.0819 | 0.0903 | 0.1385 |
| 0.10 | 0.2317 | 0.2941 | 0.3837 |
| 0.15 | 0.4436 | 0.6095 | 0.7023 |
| 0.20 | 0.6714 | 0.8789 | 0.9193 |
| 0.25 | 0.8341 | 0.9743 | 0.9821 |
| 0.30 | 0.9168 | 0.9956 | 0.9968 |

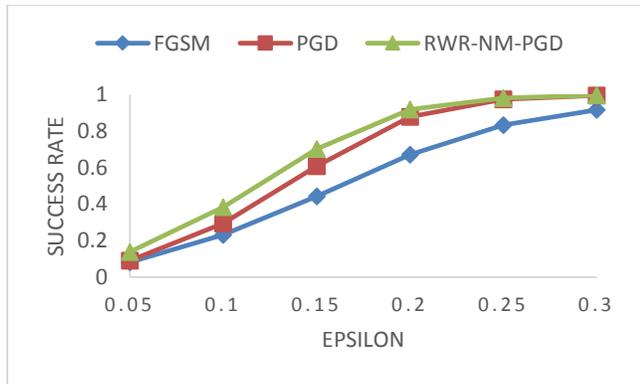

**Figure 3.** The attack results on MNIST for CNN model

**Table 2.** Attack success rate of adversarial attacks against ResNet-18 on CIFAR-10. Comparison of FGSM, PGD and RWR-NM-PGD

| CIFAR10 $\epsilon$ | Success rate | | |
|---|---|---|---|
| | FGSM | PGD | RWR-NM-PGD(Ours) |
| 0.05 | 0.6756 | 0.8919 | 0.9412 |
| 0.10 | 0.7618 | 0.9785 | 0.9981 |
| 0.15 | 0.8059 | 0.9912 | 1 |
| 0.20 | 0.8393 | 0.9942 | 1 |
| 0.25 | 0.8583 | 0.9942 | 1 |
| 0.30 | 0.8719 | 0.9942 | 1 |

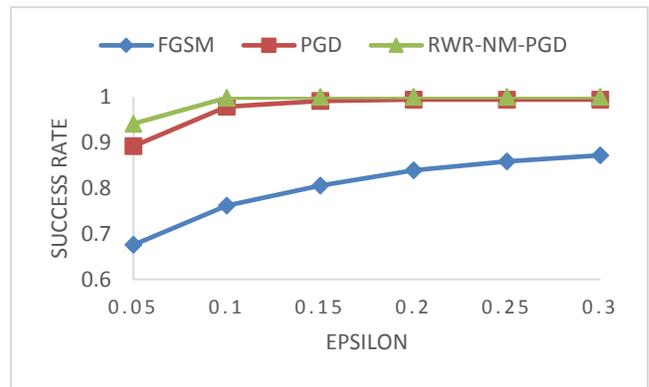

**Figure 4.** The attack results on CIFAR-10 for ResNet-18

*4.2.2 Runtime of attacks.* The runtime of generation algorithm is also an important metric to evaluate the performance of attack algorithms. The experimental results show that compared with PGD, the proposed algorithm does not increase the time cost. As shown in Table 3 and Figure 5, the runtime difference between PGD and ours is very small in the figure. When perturbations are limited to 0.15, 0.2 and 0.3, The runtime of RWR-NM-PGD is lower than or approximately equal to the runtime of PGD, and there is little difference in other cases. As shown in Table 4 and Figure 6, the runtime variation of PGD and ours are almost the same. Except for the case that the perturbations limit is 0.1, the runtime of RWR-NM-PGD algorithm is approximately equal to that of PGD in other cases. Combined with the experimental results in previous section, it is fully proved that RWR-NM-PGD with random warm restart mechanism and Nesterov momentum does not increase time cost, but improves success rate of adversarial examples attack to some extent.

**Table 3.** Runtime(s) of adversarial attacks against CNN model on MNIST. Comparison of FGSM, PGD and RWR-NM-PGD

| MNIST $\epsilon$ | RUNTIME(s) | | |
|---|---|---|---|
| | FGSM | PGD | RWR-NM-PGD(Ours) |
| 0.05 | 16.3685 | 89.2492 | 102.9069 |
| 0.10 | 16.3757 | 94.2171 | 107.0696 |
| 0.15 | 16.4194 | 86.2674 | 85.6416 |
| 0.20 | 16.3794 | 99.5298 | 80.4092 |
| 0.25 | 16.3895 | 76.8152 | 88.0087 |
| 0.30 | 16.4274 | 83.2866 | 85.9616 |

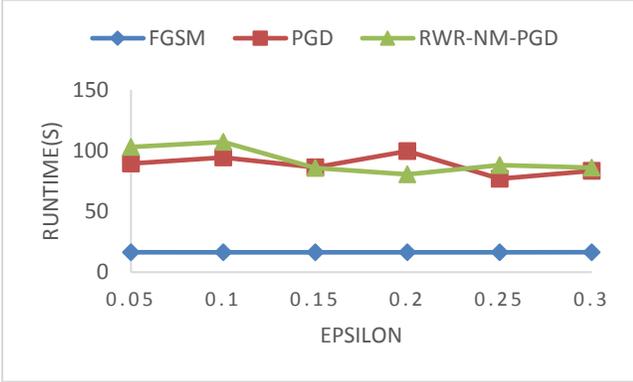

**Figure 5.** The runtime results on MNIST for CNN model

**Table 4.** Runtime(s) of adversarial attacks against ResNet-18 on CIFAR-10. Comparison of FGSM, PGD and RWR-NM-PGD

| CIFAR10 $\epsilon$ | RUNTIME(s) | | |
|---|---|---|---|
| | FGSM | PGD | RWR-NM-PGD(Ours) |
| 0.05 | 78.3162 | 266.3689 | 268.3312 |
| 0.10 | 78.1902 | 267.9883 | 277.6758 |
| 0.15 | 78.4164 | 268.9953 | 267.5608 |
| 0.20 | 78.3682 | 266.4751 | 266.3034 |
| 0.25 | 78.1593 | 267.5516 | 266.5358 |
| 0.30 | 77.8694 | 267.6211 | 266.4871 |

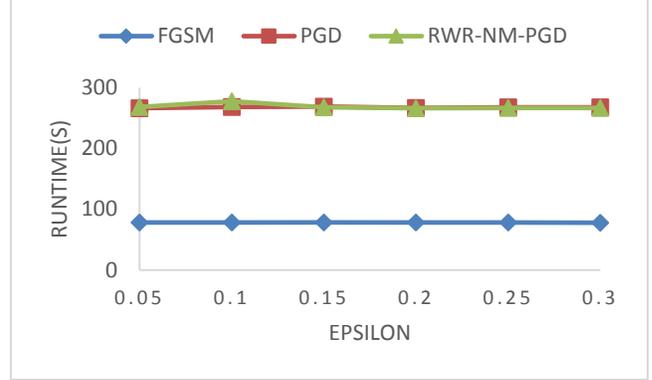

**Figure 6.** The runtime results on CIFAR-10 for ResNet-18

### 4.3 Attack defense model

There are 13 defense models, including the model based on adversarially trained and other advanced defense models. I-FGSM, PGD and RWR-NM-PGD proposed in this work are all iterative algorithms based on gradient optimization. As shown in the table, under the same experimental setting, our method has achieved the best attack success rate in all 13 defense models compared with baselines. Our method has average attack success rate of 46.3077%, which is 27.19% higher than I-FGSM and 9.27% higher than PGD.

**Table 5.** Attack success rate of adversarial attacks against defense models on CIFAR-10. Comparison of I-FGSM, PGD and RWR-NM-PGD

| Success rate | | | |
|---|---|---|---|
| Defense Models | I-FGSM | PGD | RMR-NM-PGD |
| cifar10-pgd_at | 0.440 | 0.534 | 0.540 |
| cifar10-wideresnet_trades | 0.385 | 0.434 | 0.451 |
| cifar10-feature_scatter | 0.220 | 0.281 | 0.404 |
| cifar10-robust_overfitting | 0.378 | 0.432 | 0.459 |
| cifar10-rst | 0.293 | 0.368 | 0.392 |
| cifar10-fast_at | 0.440 | 0.516 | 0.533 |
| cifar10-at_he | 0.345 | 0.382 | 0.443 |
| cifar10-pre_training | 0.360 | 0.411 | 0.432 |
| cifar10-mmc | 0.411 | 0.421 | 0.545 |
| cifar10-free_at | 0.438 | 0.548 | 0.563 |
| cifar10-awp | 0.322 | 0.359 | 0.389 |
| cifar10-hydra | 0.325 | 0.391 | 0.415 |
| cifar10-label_smoothing | 0.376 | 0.434 | 0.454 |
| Average success rate | 0.364 | 0.424 | 0.463 |

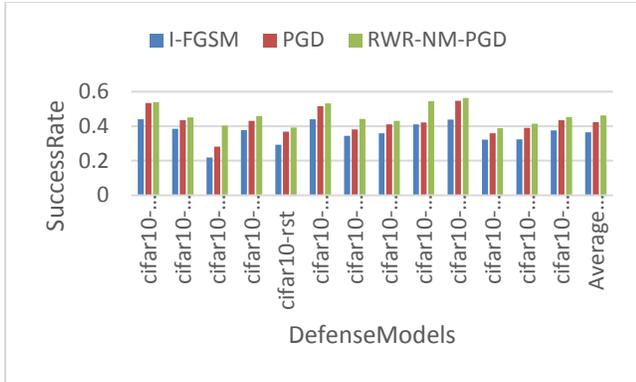

**Figure 7.** The attack results on CIFAR-10 for defense models

At last, the RWR-NM-PGD proposed in this work ranks top 1% in CVPR2021 AIC-VI: White-box Adversarial Attacks on ML Defense Models hold by Alibaba, Tsinghua University and Illinois. It fully proves the effectiveness and innovation of our method in the task of adversarial examples attacks.

## 5 CONCLUSION

In our work, based on the research and analysis of previous work, we propose two effective methods to improve adversarial attacks, improved Nesterov momentum projected gradient descent method (NM-PGD) and random warm restart method (RWR). The RWR-NM-PGD algorithm proposed in this work is inspired by the adversarial examples generation algorithm based on gradient optimization. Through the research of Nesterov momentum optimization algorithm in neural network training, the optimization process of the algorithm is improved by parameter replacement. The improved method saves the runtime cost on the basis of accelerating convergence. It also updates the optimization direction effectively by using accumulated gradient to avoid falling into local optimal solution. Experiments prove that the algorithm proposed in this work has higher attack success rate than baselines in both normal models and defense models, and has better universality and transferability.

In the future, our work will continue to research how to improve the gradient optimization method, further speed up the generation of adversarial examples and avoid falling into local optima. Meanwhile, we explore how to better design iterative attacks, reduce the number of invalid attacks, generate more effective adversarial examples, and shift the research points to adversarial examples transfer attacks.

## ACKNOWLEDGMENTS

The authors appreciate the strong support and valuable comments provided by reviewers. Authors thank the support from the adversarial robustness benchmark ARES provided by Tsinghua University. Authors are also supported form Major Program of Technological Innovation 2018AAA063. We sincerely appreciate CVPR2021 AIC-VI: White-box Adversarial Attacks on ML Defense Models.

# A  DETAILS OF THE ALGORITHMS

## A.1 Algorithm 1

**Input:** normal example $x$ with true label $y^{true}$;
classifier $f$ with loss function $L$;
**Input:** momentum decay coefficient $\mu$; step size $lr$;
maximum iterations $T$; perturbation constraint $\epsilon$.
**Output:** adversarial example $x^{adv}$.

1  $lr = \frac{\epsilon}{T}, g_0 = 0$
2  random initialize $x_0^{nm} = x + rand(x)$
3  for $t = 0$ to $T - 1$ do
4      $g = 0$
5      obtain the gradient $g$
        by $g = \nabla_x J(x_t^{nm}, y^{true})$
6      update $x_{t+1}^{nm}$ by
        $x_{t+1}^{nm} = x_t^{nm} + \mu^2 \cdot g_t - (1 + \mu) \cdot lr \cdot sign(g)$
7      clip $x_{t+1}^{nm}$ by $x_{t+1}^{nm} = proj_x^{\epsilon}\{x_{t+1}^{nm}\}$
8      update $g_{t+1}$ by $g_{t+1} = \mu \cdot g_t + \frac{g}{\|g\|_1}$
9  end
10 return $x^{adv} = x_T^{nm}$

## A.2 Algorithm 2

**Input:** classification model $f$; warm restarts $i$;
global iterations $epoch$ and maximum value $epoch_{max}$.
**Input of $i$ time restart:**
maximum step size $lr_{max}^i$ and minimum step size $lr_{min}^i$;
current step size $lr_{cur}^i$; maximum iterations of cycle $T_i$;
expansion coefficient $T_{mul}$;
iterations change from beginning to end of each restart.
**Output:** adversarial example $x^{adv}$.

1  initialize $T_0 = 2$, $i = 1$
2  for $i = 1$ to $\infty$ do
3      $i = i + 1$
4      $T_i = T_{i-1} \times T_{mul}$, set $T$ of NM-PGD to $T_i$
5      initialize $T_{cur} = 0$
6      for $T_{cur} = 0$ to $T_i$ do
7          $lr_{cur}^i = lr_{min}^i + \frac{1}{2}(lr_{max}^i - lr_{min}^i)\left(1 + cos\left(\frac{T_{cur}}{T_i}\pi\right)\right)$
8          set $lr$ of NM-PGD to $lr_{cur}^i$
9          run current iteration of NM-PGD
10         $epoch = epoch + 1$
11         if $epoch \geq epoch_{max}$ then
               finish and return $x^{adv}$
12     end
13     set input $x$ of next restart to
           the output $x^{adv}$ of current restart
14 end